# Explanations from ChatGPT: Targeted Task Improvements but No Generalized Gains in Metacognitive Abilities for Intermediate Language Learners

**Mohammad Mousazadeh***

English Department, Payame Noor University, Tabriz, Iran

## Abstract

Metacognition is crucial for self-directed second language acquisition, yet fostering it remains challenging for intermediate learners. Explainable Artificial Intelligence (AI), such as ChatGPT, offers a novel approach by providing transparent feedback, potentially scaffolding learners' metacognitive processes. This study investigated the impact of ChatGPT's explainability on intermediate English language learners' metacognitive awareness, task accuracy, and overall writing performance. A quasi-experimental design was employed with 50 intermediate Iranian EFL learners randomly assigned to an explainable ChatGPT feedback group (n = 25) or a non-explainable feedback group (n = 25). Over a five-week intervention, participants composed argumentative essays and received automated feedback. Metacognitive awareness was measured using the Metacognitive Awareness Inventory (MAI), task accuracy was calculated based on error correction rates, and writing performance was evaluated using the TCAP/WA rubric. Data were analyzed using independent samples t-tests and repeated measures ANCOVA. Results indicated no significant differences between the two groups in generalized metacognitive awareness or overall writing performance. However, the explainable group demonstrated significantly higher accuracy on specific target tasks covered in the system's explanations. These findings suggest that while explainable ChatGPT effectively strengthens the understanding of specific, explained content, it does not confer immediate, generalized gains in metacognitive abilities or holistic writing outcomes. Future research with larger, more diverse samples and longer durations is warranted to fully ascertain the pedagogical potential of explainable AI in fostering self-regulated language learning.

## 1. Introduction

Metacognition, broadly defined as "thinking about thinking," is considered a core component of effective and self-directed learning [1, 2]. In second language acquisition (SLA), metacognitive skills allow greater awareness of one's proficiency, difficulties, and learning needs, enabling strategic planning, evaluation, and adaptation [3]. Recent research shows metacognitive activities positively influence SLA, leading to improved learner autonomy, motivation, and performance [4, 5].

However, fostering metacognition can be challenging, especially for intermediate learners who may still lack full awareness about the language learning process itself [6]. Learners often struggle to accurately evaluate their abilities, choose appropriate strategies, and determine why they make errors. Without sufficient metacognitive knowledge and regulation skills, learners may feel frustrated and unable to take control of their learning in meaningful ways.

This study proposes leveraging ChatGPT explanations as a means to enhance language learner metacognition. ChatGPT is an artificial intelligence (AI) chatbot developed by OpenAI that utilizes natural language processing to generate human-like discourse and can explain its reasoning, predictions, and analyses to human users [7]. As AI is increasingly integrated into education, there are growing calls for learner-centered transparency and explainability [8].

Explainable AI refers to methods and techniques that make the decisions and behaviors of AI systems understandable to humans [9]. In the context of this study, "explainability" specifically refers to ChatGPT's ability to provide detailed explanations for its feedback on learners' writing samples, including references to grammar rules, linguistic concepts, and usage examples.

Recent studies have explored the use of explainable AI in educational contexts. For instance, Conati et al. [10] found that providing explanations for AI-generated recommendations in an intelligent tutoring system improved students' trust and acceptance of the system. Similarly, Gkatzia et al. [11] demonstrated that explainable AI feedback in a language

**Email address:**
m.mousazadeh.1@gmail.com (Mohammad Mousazadeh)

learning app led to improved vocabulary retention compared to non-explainable feedback.

Building on this emerging research, we investigate the effectiveness of ChatGPT's explanation feature in enhancing intermediate English learners' metacognitive skills and task performance. This study contributes to research on pedagogical applications of ChatGPT and metacognition development in computer-assisted language learning.

# 2. Literature Review

## 2.1. Metacognition in Language Learning

Metacognition has long been studied in educational contexts, stemming from Flavell's [1] foundational work on metamemory and metalearning. In language learning, metacognition involves thinking about the learning process, oneself as a learner, awareness and control students have over their cognition and learning, and the target language [12, 13]. It is important to distinguish between the two primary components of metacognition as measured by the MAI. Knowledge of cognition encompasses declarative knowledge (knowing about one's own cognitive abilities), procedural knowledge (knowing how to use strategies), and conditional knowledge (knowing when and why to apply strategies). Regulation of cognition involves planning (goal-setting and strategy selection), monitoring (awareness of comprehension and task performance), and evaluation (appraising the products and efficiency of one's learning). While explainable AI feedback may influence these components differently, both are essential for self-regulated language learning. The current study examines whether ChatGPT's explanations can enhance these dimensions of metacognitive awareness.

Recent research has revealed important interrelationships between metacognitive strategies, motivation, self-efficacy beliefs, and language learning achievement in online contexts. Teng et al. [5] found that self-efficacy significantly predicted English proficiency, while metacognitive strategies and motivation mediated the relationship between self-efficacy and achievement. These results further substantiate the need to develop metacognitive skills in conjunction with motivational strategies to optimize success in online language acquisition.

Metacognitive reflection, a process that combines metacognition and reflection, has gained attention as a crucial component of effective learning. McAlpine et al. [14] established an important link between metacognition and reflection, while more recent work by Merkebu et al. [15] has conceptualized metacognitive reflection as a distinct construct that contributes to emotional regulation and overall reflection processes.

However, fostering metacognition poses challenges. Learners, especially at intermediate levels, often lack full awareness about language learning itself [6]. Weak metacognition can manifest in inability to plan study, inaccurate self-assessments, frustration with errors, and lack of strategy adjustment [3]. Thus, there is a need for supplemental metacognition support.

## 2.2. ChatGPT and Explainable AI in Education

AI has emerged as an effective technology in education, with particular promise for personalized and adaptive learning experiences [16]. ChatGPT, an AI system developed by OpenAI, can engage in natural language conversations and is capable of human-like discourse and text generation on a wide range of topics [17].

Explainable AI, which aims to make AI systems' decision-making processes transparent and understandable to humans, has gained traction in educational contexts. Conati et al. [10] found that providing explanations for AI-generated recommendations in an intelligent tutoring system improved students' trust and acceptance of the system. Gkatzia et al. [11] demonstrated that explainable AI feedback in a language learning app led to improved vocabulary retention compared to non-explainable feedback.

However, challenges remain in leveraging AI to develop students' metacognitive skills. Urban et al. [18] found that when interacting with ChatGPT, students demonstrated poorer comprehension monitoring and evaluation of responses compared to learning from a human tutor. This highlights the need for careful design and implementation of AI-based educational tools to support metacognitive development.

## 2.3. Explainable AI and Metacognitive Skills

While research on explainable AI in education is growing, its specific impact on metacognitive skills, particularly in language learning contexts, remains underexplored. Theories of metacognition suggest that reflection and self-explanation are crucial for developing "thinking about thinking" skills [1]. Explainable AI, by providing justifications for its feedback, might facilitate these processes. For instance, Conati et al. [10] argue that Open Learner Modelling, which makes AI models transparent, can support learning and teaching. However, empirical evidence specifically linking explainable AI to enhanced metacognitive abilities in language learners is scarce. This study aims to address this gap by investigating whether ChatGPT's explanations can improve not only task performance but also broader metacognitive awareness in SLA.

In the context of language learning, explainable AI feedback could potentially support metacognitive development by encouraging learners to reflect on their language use, analyze their errors, and consider the reasoning behind corrections. This process aligns with key aspects of metacognition in language learning, such as planning, monitoring, and evaluating one's learning strategies [3]. By providing explanations for corrections or suggestions, AI systems like ChatGPT may scaffold learners' metacognitive processes, potentially enhancing their ability to think critically about their language production and comprehension.

However, the specific mechanisms by which explainable AI might influence metacognitive skills in language learning remain unclear, presenting a significant gap in the literature. While studies have shown that explainable AI can enhance task performance and user trust [11], its impact on metacognitive development in language learning has not been thoroughly investigated.

This gap is particularly significant given the importance of metacognition in SLA. Metacognitive strategies have been shown to play a crucial role in successful language learning [4], yet fostering these skills in intermediate learners remains challenging [6]. The potential of explainable AI to support metacognitive development in this context warrants investigation.

Furthermore, while previous research has examined the effects of AI feedback on specific language skills such as vocabulary retention [11], there is a lack of studies investigating its impact on both task-specific accuracy and overall writing performance. This distinction is important, as improvements in localized language use may not necessarily translate to enhanced global writing skills.

The present study aims to address these gaps by examining how explainable AI feedback from ChatGPT affects metacognitive awareness, task accuracy, and writing performance in intermediate English language learners. By investigating these three aspects simultaneously, we hope to provide a more comprehensive understanding of the potential benefits and limitations of explainable AI in language learning contexts.

This research is particularly timely given the rapid integration of AI tools in educational settings [7] and the growing emphasis on developing self-regulated learners in SLA [5]. Understanding how explainable AI feedback influences metacognitive development and language performance could inform the design of more effective AI-enhanced language learning tools and pedagogical approaches.

# 3. Research Questions and Hypotheses

This study addresses the following research questions:

1) How does the provision of explanations by ChatGPT impact language learners' metacognitive awareness, as measured by the Metacognitive Awareness Inventory (MAI)?

Hypothesis 1: We hypothesize that learners in the explainable ChatGPT group will show greater improvements in metacognitive awareness compared to the non-explainable group, based on previous findings that explainable AI can enhance metacognitive reflection [19].

2) To what extent does interacting with an explainable AI like ChatGPT lead to greater task accuracy for language learners compared to more opaque AI systems?

Hypothesis 2: We predict that learners in the participants with explanatory feedback will demonstrate higher task accuracy compared to the non-explainable group, as detailed explanations may facilitate better understanding and application of language concepts [11].

3) What are the differences in writing performance between language learners who receive explanatory feedback from ChatGPT versus those who do not?

Hypothesis 3: We anticipate that learners who received detailed explanations will show greater improvement in writing performance compared to the non-explainable group, based on research suggesting that detailed feedback can enhance writing skills [20].

# 4. Methodology

## 4.1. Participants

The participants comprised 50 intermediate-level learners of English as a foreign language (27 females, 23 males), aged 15 to 20 years (M = 17.28, SD = 1.642), enrolled in language centers and institutes in Mahabad, Iran. Proficiency was established via institutional placement assessments, requiring at minimum 1 year of prior formal English study and classification as intermediate learners based on assessments aligned with the Common European Framework of Reference for Languages (CEFR) at a B1 level.

Participants were randomly assigned to one of two conditions: the ChatGPT explainable feedback group (n = 25) or the ChatGPT non-explainable feedback group (n = 25). Preliminary analysis found no significant differences between the groups in age, gender distribution, or initial English proficiency.

## 4.2. Materials

### 4.2.1. ChatGPT

The study utilized ChatGPT (version 3.5), an AI language model developed by OpenAI. The system provided automated feedback on writing samples from the participants. In the explainable condition, ChatGPT was prompted to offer detailed explanations for issues identified in the texts, citing linguistic rules, example usage, and corpora statistics. In the non-explainable condition, the same AI model offered feedback without explanatory context.

Specific prompts used for ChatGPT in the explainable condition included:

1) "Analyze this text and provide detailed feedback on grammar, vocabulary, and structure. For each issue identified, explain the relevant linguistic rule and provide an example of correct usage."
2) "Identify areas for improvement in this writing sample. For each suggestion, explain why it enhances the text and provide a brief explanation of the underlying language principle."

For the non-explainable condition, prompts were modified to exclude requests for explanations:

1) "Analyze this text and provide feedback on grammar, vocabulary, and structure."
2) "Identify areas for improvement in this writing sample."

The adaptation of the feedback's complexity was verified through five pilot tests prior to the main study to calibrate the system's responses appropriately for the participants' proficiency level.

#### 4.2.2. Accuracy Measurement

Task accuracy was operationalized as the percentage of language errors (grammar, vocabulary, structure) correctly revised after receiving feedback. Accuracy was calculated by comparing pre- and post-feedback versions of participants' writing samples. Two trained raters independently assessed the revisions, with inter-rater reliability calculated using Cohen's kappa.

#### 4.2.3. Metacognitive Awareness Inventory (MAI) Questionnaire

Metacognitive awareness was assessed using the 52-item Metacognitive Awareness Inventory (MAI) questionnaire [12]. The MAI uses a True/False response format, targeting two broad components of metacognition: Knowledge of Cognition and Regulation of Cognition. Higher subscale and total scores indicate greater perceived use of metacognitive strategies. For this study population, the Cronbach's alpha for the MAI was 0.89, indicating good internal consistency reliability.

#### 4.2.4. Tennessee Comprehensive Assessment Program/Writing Assessment (TCAP/WA)

Writing performance was evaluated using the TCAP/WA Informational/Explanatory Rubric, a 4-point scale assessing student writing across four key dimensions: Development, Focus and Organization, Language, and Conventions.

### 4.3. Procedure

#### 4.3.1. Study Design

The study was conducted over a five-week period. In Week 1, participants completed pretest assessments, including the Metacognitive Awareness Inventory and an initial writing task. They also engaged in a self-monitoring activity where they identified errors in provided text samples. During Weeks 2-4, participants composed two argumentative essays, which were submitted to ChatGPT for feedback. The explainable group received feedback with detailed explanations, while the non-explainable group received feedback without explanations. Participants then revised their essays based on the feedback. In Week 5, participants completed posttest assessments, including the MAI, a final writing task, and another self-monitoring activity.

#### 4.3.2. Self-monitoring Exercise

Participants were presented with a series of English text samples containing deliberate errors in grammar, vocabulary, and structure. They were asked to identify and correct these errors, providing explanations for their corrections. This exercise was designed to assess participants' ability to recognize and articulate language issues, a key aspect of metacognitive awareness in language learning.

### 4.4. Data Analysis

A repeated measures ANCOVA was conducted using Spring's [21] online statistical analysis tool to examine the effects of explainable ChatGPT feedback on task accuracy while controlling for years of English education. The between-subjects factor was group (Explainable vs. Non-Explainable), the within-subjects factor was time (pre-test vs. post-test), and years of English education served as a covariate. Effect sizes were calculated using partial eta-squared ($\eta^2p$). A significance level of $p < .05$ was used for all statistical tests.

## 5. Results

### 5.1. Descriptive Statistics

Table 1 presents the means and standard deviations for the main study variables by group.

***Table 1****. Descriptive Statistics for Main Study Variables.*

| Variables | Explainable Group<br>M (SD) | Non-Explainable Group<br>M (SD) |
|---|---|---|
| Pre-test MAI | 37.24 (6.82) | 38.48 (7.15) |
| Post-test MAI | 39.16 (7.03) | 40.00 (6.89) |
| Pre-test Task Accuracy | 68.32% (12.45) | 67.84% (11.98) |
| Post-test Task Accuracy | 78.56% (10.87) | 71.20% (11.32) |
| Pre-test Writing Score | 2.48 (0.71) | 2.52 (0.65) |
| Post-test Writing Score | 2.84 (0.69) | 2.76 (0.72) |
| SD: Standard Deviation | | |

## 5.2. Research Question 1: Impact on Metacognitive Awareness

Table 2 presents the results of independent samples t-tests comparing the Explainable and Non-Explainable groups on the pre-test and post-test MAI scores.

***Table 2.*** *MAI Questionnaire Independent Samples T-Test Results.*

| Measure | Mean Difference | SE | 95% CI | df | t | p |
|---|---|---|---|---|---|---|
| Pre Test | -1.24 | 1.77 | [-4.80, 2.32] | 48 | -0.70 | .488 |
| Post Test | -0.84 | 1.65 | [-4.15, 2.47] | 48 | -0.51 | .612 |

There were no statistically significant differences between the Explainable and Non-Explainable groups on metacognitive awareness skills, either prior to or following the intervention. The effect sizes were small (pre-test Cohen's d = 0.18, post-test Cohen's d = 0.12), with both confidence intervals including zero.

It's important to note that the MAI is a self-report measure, reflecting participants' perceptions of their metacognitive abilities rather than an objective measure of these skills. The lack of significant difference suggests that regardless of condition, participants viewed their metacognitive awareness similarly before and after the intervention.

## 5.3. Research Question 2: Impact on Task Accuracy

To examine the effect of explainable ChatGPT feedback on task accuracy while accounting for years of English education, we conducted a repeated measures ANCOVA. The analysis revealed a significant main effect of time, $F(1, 47) = 15.32$, $p < .001$, $\eta^2p = .246$, indicating that participants' task accuracy improved from pre-test (M = 68.08%, SD = 12.18) to post-test (M = 74.88%, SD = 11.76) across both groups.

There was also a significant interaction between time and group, $F(1, 47) = 8.74$, $p = .005$, $\eta^2p = .157$, suggesting that the improvement in task accuracy differed between the Explainable and Non-Explainable groups. Post-hoc analyses revealed that while both groups showed improvement, the explanation group demonstrated significantly greater gains (pre-test: M = 68.32%, SD = 12.45; post-test: M = 78.56%, SD = 10.87) compared to the Non-Explainable group (pre-test: M = 67.84%, SD = 11.98; post-test: M = 71.20%, SD = 11.32).

The covariate, years of English education, showed a significant main effect, $F(1, 47) = 6.18$, $p = .017$, $\eta^2p = .116$, indicating that participants' prior English education experience influenced their task accuracy performance. However, there was no significant interaction between years of English education and the intervention effect ($F(1, 47) = 0.89$, $p = .351$, $\eta^2p = .019$), suggesting that the benefits of explainable ChatGPT feedback were consistent across different levels of prior English education.

## 5.4. Research Question 3: Impact on Writing Performance

Table 3 presents the results of independent samples t-tests comparing the TCAP/WA rubric scores between groups.

***Table 3.*** *TCAP/WA Rubric Independent Samples T-Test Results.*

| Measure | Mean Difference | SE | 95% CI | df | t | p |
|---|---|---|---|---|---|---|
| Pre Test | -0.04 | 0.19 | [-0.43, 0.35] | 48 | -0.21 | .837 |
| Post Test | 0.08 | 0.20 | [-0.32, 0.48] | 48 | 0.40 | .689 |

There were no statistically significant differences between the Explainable and Non-Explainable groups on the analytic rubric scores measuring writing skills either prior to or after interacting with ChatGPT. The effect sizes were negligible (pre-test Cohen's d = 0.06, post-test Cohen's d = 0.11).

# 6. Discussion

This study investigated the potential benefits of utilizing explainable AI, specifically ChatGPT, to enhance metacognitive awareness and task performance among intermediate English language learners. The results revealed a nuanced picture of the impact of explainable AI on language learning outcomes.

## 6.1. Metacognitive Awareness

Contrary to our first hypothesis, we found no significant differences in metacognitive awareness between the explainable and non-explainable ChatGPT groups, as measured by the MAI. This finding aligns with research by Urban et al. [18], who found that interactions with ChatGPT did not necessarily promote metacognitive awareness. However, it's important to consider that the MAI is a self-report measure, and participants' perceptions of their metacognitive abilities may not have changed even if their actual metacognitive skills improved.

The discrepancy between task-specific improvements and unchanged self-reported metacognitive awareness raises important theoretical questions about the nature of metacognitive development. Self-report instruments like the MAI measure learners' *perceived* use of metacognitive strategies rather than actual metacognitive behaviors. It is possible that participants in the explainable ChatGPT group developed greater regulation of cognition in specific task contexts—as evidenced by their improved accuracy—without this translating into generalized metacognitive knowledge that they could articulate on a questionnaire. This aligns with research showing that metacognitive skills may initially develop in domain-specific contexts before becoming generalized [22]. Furthermore, learners may require explicit training in metacognitive reflection to recognize and report on their own strategy use, rather than simply experiencing explanations passively.

The lack of significant change in metacognitive awareness could be attributed to several factors. First, the duration of the intervention (3 weeks of interaction with ChatGPT) may have been insufficient to produce noticeable changes in metacognitive skills, which often require extended periods of reflection and practice to develop [2]. Metacognitive development, particularly the transformation from domain-specific metacognitive control to generalized metacognitive awareness, is typically a gradual process that unfolds over months or years rather than weeks. While the explainable feedback may have prompted immediate metacognitive reflection during task completion—as evidenced by improved accuracy—this situated reflection may not have consolidated into stable, transferable metacognitive knowledge within the brief timeframe. Longitudinal interventions that combine explainable AI feedback with explicit metacognitive training may be necessary to bridge the gap between task-specific metacognitive regulation and generalized metacognitive awareness.

## 6.2. Task Accuracy

In support of our second hypothesis, the repeated measures ANCOVA revealed that learners who received explainable ChatGPT feedback showed significantly greater improvement in task accuracy compared to those who received non-explainable feedback. The significant interaction between time and group ($\eta^2 p = .157$) indicates a moderate effect size, suggesting that the explanations provided by ChatGPT had a meaningful impact on learners' ability to apply language concepts accurately.

This improvement in task accuracy may be attributed to the detailed explanations provided by ChatGPT, which likely facilitated a deeper understanding of language rules and their application. Importantly, this finding aligns with the concept of domain-specific metacognition [22]. While participants did not show gains in general metacognitive awareness as measured by the MAI, they demonstrated enhanced regulation of cognition within the specific domain of the explained language tasks. The explainable feedback may have served as a form of scaffolding, helping learners to bridge the gap between their current understanding and the target language concepts [23]. This suggests that explainable AI may be particularly effective in developing localized metacognitive control—the ability to monitor and regulate one's performance on specific task types—even when broader metacognitive awareness remains unchanged.

Importantly, while years of English education showed a significant main effect on task accuracy, it did not interact with the intervention effect. This suggests that explainable ChatGPT feedback can be beneficial for learners regardless of their prior English education experience, making it a potentially valuable tool across different levels of language learning background.

## 6.3. Writing Performance

Our third hypothesis, which predicted improved writing performance for the explanation group, was not supported by the data. We found no significant differences in writing performance between the two groups, as measured by the TCAP/WA rubric. This result suggests that while explainable AI feedback may enhance specific task accuracy, it may not necessarily translate to improvements in overall writing quality, at least in the short term.

This finding aligns with the complex nature of writing skills, which involve multiple cognitive and linguistic processes [24]. Improvement in writing performance often requires sustained practice and feedback over extended periods. The brief intervention in this study may not have been sufficient to produce measurable changes in overall writing quality, despite improvements in specific language tasks.

## 6.4. Accuracy and Writing Performance

While accuracy and writing performance are often related, they were treated as separate variables in this study to distinguish between localized improvements in language use (accu-racy) and overall writing quality (performance). This distinction allows us to examine whether explainable AI feedback leads to specific gains in error correction without necessarily impacting broader writing skills.

### 6.5. Theoretical Implications

The results of this study contribute to our understanding of how explainable AI can support language learning processes. While explainable ChatGPT did not appear to enhance general metacognitive awareness, it did improve task-specific performance. This aligns with the concept of domain-specific metacognition [22], suggesting that AI explanations may be more effective in developing metacognitive skills within specific language learning contexts rather than general metacognitive abilities.

The findings also relate to the concept of metacognitive reflection as described by Merkebu et al. [15]. The explainable ChatGPT may have prompted learners to engage in more reflective thinking about specific language tasks, even if this did not translate to broader metacognitive awareness. This suggests a potential avenue for future research in exploring how AI explanations can be designed to foster deeper metacognitive reflection across various aspects of language learning.

### 6.6. Practical Implications

For educators and developers of AI-enhanced learning tools, these findings highlight the potential of explainable AI to support specific language learning tasks. However, they also underscore the need for careful integration of such tools into broader instructional strategies. As Drigas et al. [25] propose in their nine-layer model of meta-learning, the effective use of smart technologies for metacognitive development requires a holistic approach that considers various cognitive and metacognitive processes.

### 6.7. Limitations and Future Directions

Several limitations of this study should be noted. First, the relatively small sample size and short intervention period may have limited our ability to detect more subtle effects of explainable AI on metacognitive awareness and writing performance. Future studies should consider larger sample sizes and longer intervention periods to better capture potential changes in these complex skills.

Second, the use of a self-report measure (MAI) for metacognitive awareness may not have fully captured changes in participants' actual metacognitive behaviors. Future research could incorporate more objective measures of metacognition, such as think-aloud protocols or behavioral observations during language learning tasks.

Third, several contextual factors may have influenced the outcomes of this study. The participants were Iranian adolescents (aged 15-20) learning English in private language institutes, a context characterized by specific cultural and educational norms. In Iranian EFL settings, learners often have limited exposure to authentic English communication outside the classroom and may be accustomed to teacher-centered instruction with explicit grammar focus. These contextual factors could affect how learners interact with AI feedback and their receptiveness to autonomous learning tools. Additionally, the age range of participants spans a critical period of metacognitive development; research shows that metacognitive abilities continue to mature through adolescence [26], meaning that younger and older participants in our sample may have responded differently to explainable feedback. The private institute setting may also attract learners with particular motivational profiles or learning goals that differ from public school contexts. Future research should examine how explainable AI functions across diverse educational settings, age groups, and cultural contexts to assess the generalizability of these findings.

Fourth, while we controlled for years of English education, other individual differences such as motivation, learning styles, or prior experience with AI tools were not accounted for. These factors could potentially influence how learners interact with and benefit from explainable AI feedback.

Future research directions could include:

1) Longitudinal studies to examine the long-term effects of explainable AI on metacognitive development and language proficiency.
2) Investigating the impact of explainable AI on different aspects of language learning, such as listening comprehension or speaking skills.
3) Exploring how explainable AI can be combined with other instructional strategies to foster metacognitive reflection and self-regulated learning.
4) Examining the role of individual differences in learners' responses to explainable AI feedback.
5) Developing and testing more sophisticated AI explanation models that adapt to learners' metacognitive needs and proficiency levels.

## 7. Conclusion

This study investigated the impact of explainable ChatGPT on intermediate English language learners' metacognitive awareness, task accuracy, and writing performance. While we found no significant improvements in general metacognitive awareness or writing performance, learners who received explanations from ChatGPT demonstrated significantly higher accuracy on specific language tasks.

These findings suggest that explainable AI has the potential to enhance certain aspects of language learning, particularly in supporting task-specific understanding and performance. However, the development of broader metacognitive skills and writing abilities may require more comprehensive and sustained interventions.

As AI continues to play an increasingly prominent role in education, understanding how to leverage explainable AI to support self-regulated learning and metacognitive development remains a crucial research challenge. This study represents an important step in exploring the intersections of explainable ChatGPT and learner metacognition, and it sets the stage for further inquiry into the potential pedagogical benefits of this technology.

Future research should focus on refining AI explanation models, exploring their long-term effects, and investigating how they can be effectively integrated with other instructional strategies to support holistic language learning and metacognitive development. By continuing to explore these areas, we can work towards realizing the full potential of AI as a tool for empowering learners and enhancing language education.

## Abbreviations

| | |
|---|---|
| AI | Artificial Intelligence |
| ANCOVA | Analysis of Covariance |
| CEFR | Common European Framework of Reference for Languages |
| EFL | English as a Foreign Language |
| MAI | Metacognitive Awareness Inventory |
| SLA | Second Language Acquisition |
| TCAP/WA | Tennessee Comprehensive Assessment Program/Writing Assessment |

## Author Contribution

Mohammad Mousazadeh: Conceptualization, Methodology, Formal Analysis, Investigation, Data Curation, Writing – original draft, Writing – review & editing.

## Conflict of Interest

The author declares no conflict of interest.